# One-Chip Solution to Intelligent Robot Control: Implementing Hexapod Subsumption Architecture Using a Contemporary Microprocessor

**Nikita Pashenkov & Ryuichi Iwamasa**
GK Tech Inc.,
3-30-14, Takada, Toshima-ku, Tokyo, 171-0033, Japan
e-mail: nik@media.mit.edu, iwamasa@gk-design.co.jp

*Abstract: This paper introduces a six-legged autonomous robot managed by a single controller and a software core modeled on subsumption architecture. We begin by discussing the features and capabilities of IsoPod, a new processor for robotics which has enabled a streamlined implementation of our project. We argue that this processor offers a unique set of hardware and software features, making it a practical development platform for robotics in general and for subsumption-based control architectures in particular. Next, we summarize original ideas on subsumption architecture implementation for a six-legged robot, as presented by its inventor Rodney Brooks in 1980's. A comparison is then made to a more recent example of a hexapod control architecture based on subsumption. The merits of both systems are analyzed and a new subsumption architecture layout is formulated as a response. We conclude with some remarks regarding the development of this project as a hint at new potentials for intelligent robot design, opened up by a recent development in embedded controller market.*

## 1. Introduction

As a preliminary application area for the technology platform of a robotics project, we have recently undertaken a task of implementing a controller architecture for an autonomous six-legged robot. The project was spurred in part by the availability of a new embedded development board for robotics, called *IsoPod*, designed around a speedy general-purpose DSP core and marketed by a small company in Texas, USA.

As the next section will attempt to show, the combined feature set of the new processor makes it a uniquely convenient platform for the development of software controllers from the simplest to very complex machines. IsoPod's on-chip operating system in particular is designed to encourage a modular approach to software architecture design from bottom up. As a complement to this approach, we decided to implement a version of subsumption architecture, popularized by Rodney A. Brooks at MIT in 1980's , as our software controller.

This paper discusses a short overview of two subsumption architecture layouts for a six-legged robot. The first scheme is Brooks' original architecture for his seminal Gehghis hexapod. (Brooks, R. & Flynn, A., 1989) The second is a recent example of a subsumption architecture developed as a response to and an improvement upon Brooks' in terms of clarity. Our own approach led us to a new modification of hexapod subsumption which draws on both of the previous examples.

In conclusion, we discuss the details of our subsumption controller implementation as it relates to the specifics of our chosen development platform. We believe that the appearance of advanced yet inexpensive robotics platforms, such as IsoPod and its operating system *IsoMax*, open up new possibilities for the construction of intelligent robotics platforms by professionals and hobbyists alike.

## 2. IsoPod Hardware

IsoPod™ is a controller board designed with robotics applications in mind by New Micros, Inc. of Dallas, Texas. Introduced on June 2002, Isopod is marketed by the company as a complete interactively programmable system, featuring a built-in high-level language and a "virtually-parallel processing" operating system, IsoMax™. The IsoPod hardware is based around a Motorola DSP56F800-series 16-bit Digital Signal Process (DSP) processor, which runs at 80Mhz and provides a wealth of hardware features including 16 General Purpose Digital I/O lines, 2 serial channels, RS-232 and RS-422/485, CAN BUS, an SPI Interface, 8 channels of 12-bit A/D, 8 General Purpose Timers, and 12 PWM outputs. Notably, the multitude of useful features are brought out onto easy-to-access connectors that are densely populated on the 1.2" x 3.0" IsoPod board. An additional small connector board is provided so that up to 22 servos could be connected to the IsoPod. According to New Micros, the overall list of features makes IsoPod hardware suitable for dedicated control of DC motors, BDCM, stepper motors, solenoids, and other bipolar outputs. (New Micros, Inc., 2004).

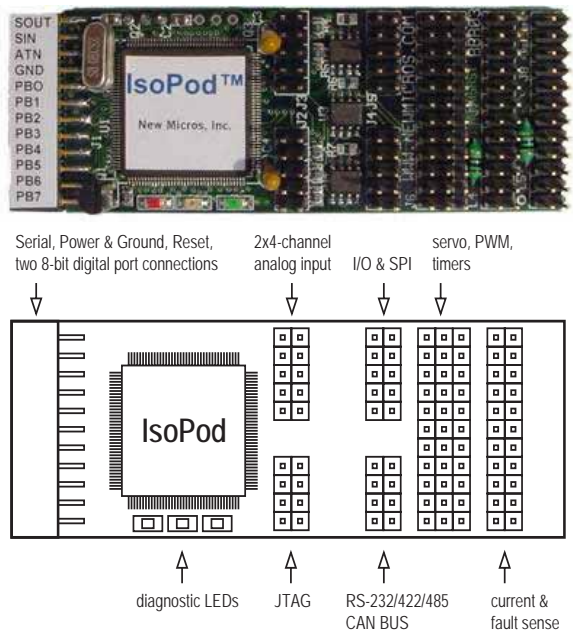

Fig. 1. IsoPod board and functional diagram.

### 3. IsoPod Software

In addition to a solid set of hardware features provided by its powerful processor, IsoPod boasts an innovative software foundation. Each IsoPod ships pre-loaded with a resident programming language IsoMax, which is based on the notion of Finite State Machines (FSM) and a procedural language derived from Forth. The key feature of IsoMax is the concept of Virtually Parallel Machine Architecture, billed by New Micros as a "new programming paradigm." VPMA allows independent virtual machines to be constructed in software, tested one by one and added seamlessly together, then run in a virtually parallel fashion. According to the maker, "The VPMA structure may hold advances in Neural Net Processing simulation, and AI applications... robots using this new paradigm may appear more 'thinking' than their predecessors."

### 4. IsoPod Programming

The development of user programs for the IsoPod is accomplished via serial communication with the board through RS-232 connection and a terminal program on a PC. The programming process consists of loading segments of programs into IsoPod RAM or Flash memory, then testing them interactively by invoking appropriate commands.

Programming real time tasks in IsoMax consists of describing virtual machines that will sense conditions, take actions, and move to new states. IsoMax provides a convenient syntax to create new processor tasks, each being a state machine, and the means of changing states in a machine. A simple virtual machine can be constructed with a few lines of code, then tested by calling up the machine state and 'installing' the machine individually or into a 'chain' of (theoretically unlimited) number of machines. The table in Fig. 2 outlines the specifics of IsoMax syntax as it relates to this process.

Between the standard clauses of IsoMax, procedural statements modeled after Forth prog-ramming language can be inserted. These statements complement IsoMax to provide a rich syntax, the structure of which resembles something of a cross between assembly code and a high-level language. Like the latter, the syntax of IsoMax is loose in regard to spacing and indentation. The concept of virtually parallel machines also encourages a programming style more akin to the structure of more recent object-oriented languages, in comparison to procedural languages that are in fact closer contemporaries of Forth.

| Step | Programming Action | Syntax |
|---|---|---|
| 1 | Name a state machine | MACHINE <name> |
| 2 | Select state | ON-MACHINE <name> |
| 3 | Name appended states | APPEND-STATE <name> |
| 4 | Describe state transitions: | |
|  | starting state | IN-STATE <state> |
|  | condition necessary to leave | CONDITION <boolean> |
|  | action to take place | CAUSES <action> |
|  | state to go to next time | THEN-STATE <state> |
|  | closing statement | TO-HAPPEN |
| 5 | Set state | <state> SET-STATE |
| 6 | Run machine | INSTALL <machine> |

Fig. 2. IsoMax state machine syntax.

### 5. IsoPod Evaluation

In our tests so far, IsoPod has proven to be a mature and dependable product, although an occasional bug or lack of documented feature reveals that IsoMax operating system is still in beta release (version 0.6 as of this writing). Of particular interest to this project are two main features of the Isopod package. Firstly, the fact that a single IsoPod controller is able to control up to 22 servos, a number suitable to construct a six-legged robot with 3 degrees of freedom per leg. Secondly, the fact that IsoPod software model is based on the concept of Finite State Machines and allows for easy construction and testing of a multitude of such modules, which can run in a virtually parallel fashion. This feature makes IsoPod a near-ideal testing bed for robot control systems based on the idea subsumption architecture, described in more detail below.

## 6. Subsumption Architecture

Subsumption architecture was developed by Rodney A. Brooks and his students at MIT Artificial Intelligence Laboratory in as the cornerstone of their "Nouvelle AI" philosophy. (Brooks, R., 1986) The basic framework of subsumption is a modular network of Finite State Machines, which can be broken into layers of control allowing the robot to operate at increasing levels of 'competence.' The control layers are made up of asynchronous modules that communicate with each other by passing messages.

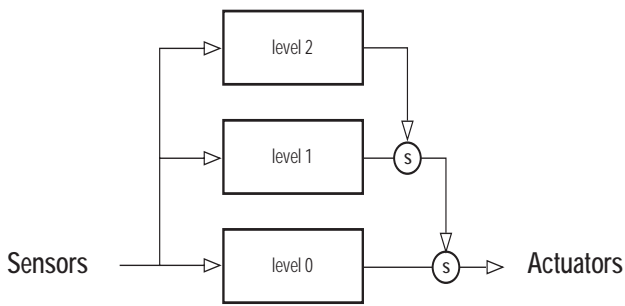

Fig. 3. Generalized subsumption diagram.

The term 'subsumption' derives from the idea that higher levels can subsume the lower levels by suppressing their outputs as needed. The advantage of this system is that a robot control systems can be built incrementally, as lower levels continue to function while higher ones are added on. The result, according to Brooks, is a development framework for robotics that offers a large number of advantages concerning robustness, buildability, and testability over the approaches emplyed in traditional or classical AI. (Brooks, R., 1990)

## 6. Hexapod Subsumption Architecture by Brooks

The earliest description of subsumption architecture for a hexapod robot is provided by Brooks in MIT A.I. Laboratory Memo from 1989. Brooks gives a clear analysis of the architecture layout in two parts. The first part describes the lower portion of the entire architecture excluding sensor feedback and some high-level behavior modules. As the principle of subsumption dictates, the lower portion of the diagram can function on its own to provide a self-sufficient behavior, called 'simple walk' in this case. (Brooks, R., 1989)

*Simple walk* consists of 36 machines which together enable a six-legged robot to walk across flat terrain. The lowest level of competence in this set consists of *alpha position* and *beta position* modules, which receive variables controlling the positions of corresponding leg motors or servos. When the system is first powered up, these machines are sent initial values that cause the robot to stand up.

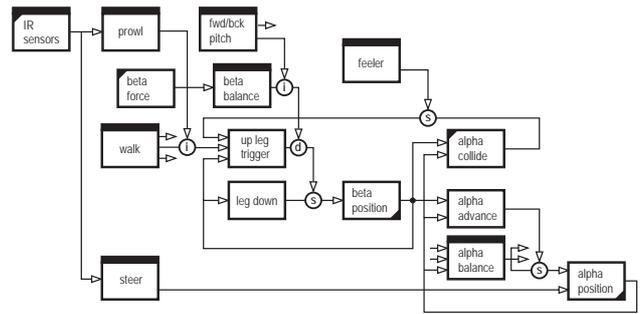

Fig. 4. Hexapod Subsumption by Brooks.

The addition of a few more machines allow the robot to make a step. The *leg down* module for each leg always tries to set the corresponding limb to downward position by writing to the *beta position* module. This message can be suppressed by the *up leg trigger* machine, causing the robot to raise the leg when activated. Once the output of *beta position* matches the desired height, *alpha advance* machine is triggered, causing the robot's leg to swing forward. The global *alpha balance* machine sums the *alpha position* outputs of each leg and writes a small counterbalancing value to inputs unsuppressed by a-advance. As a result, whenever a robot leg moves forward, the other legs move slightly back (or vice versa) making a balanced step.

Finally, walking is achieved with the addition of *walk* machine, which sends an appropriate pattern of *up leg trigger* messages that propagate according to the sequence described above. It is possible to make the robot produce a number of walking patterns, from ripple gate (one leg at a time) to tripod gate (three legs at a time), by changing the pattern stored in *walk*, without adjusting any of the modules below.

The diagram in Fig. 4 represents the complete hexapod subsumption architecture designed by Brooks. The extra modules added on top of *simple walk* allow the robot to sense its environment and to be able to traverse uneven terrain. In order for the robot to sense contact with the ground, *beta force* module is provided. A global *beta balance* module corrects the vertical displacement of all legs so that the robot maintains a stable position. On the horizontal plane, *alpha collide* module monitors any lateral forces on robot's legs that may come from obstacles, causing the legs to lift to overcome them. Since the robot is outfitted with whiskers, *feeler* module is added which helps the robot to anticipate obstacles lying ahead. Finally, the high-level *prowl* behavior module is added which receives input from *IR sensor* module. This unit is programmed such that when motion is detected by the sensors, the walking behavior is activated causing the robot to wander.

The full-featured Brooks' subsumption architecture as described above proved itself successful in real-world tests on robots built at MIT. However, a quick look at the diagram reveals a rather complicated patchwork of modules that has lost some of the elegance of the original *simple walk*. A more recent example of hexapod subsumption architecture tackles this issue in an attempt to provide a more clear and easy-to-read diagram.

## 7. Contemporary Example of Hexapod Subsumption Architecture

In 2000, Enric Celaya and Josep M. Porta of Institut de Cibernètica, Barcelona, developed a controller for a six-legged robot that allows it to walk on rough terrain using force feedback. (Celaya, E. & Porta, J., 2000) The researchers suggest that their initial approach was to add a force-compliance layer on top of a subsumption-based controller tailored to walk on flat surfaces and thus build the controller in a purely incremental way. However, Porta & Celaya found that they needed to make substantial modications to the existing layers first.

For example, the previously described *walk* and *alpha advance* behaviors had to be redesigned or completely eliminated. In addition, modularity of the architecture was lost as layered structure had to be defined differently for compliant (force-feedback) and non-compliant (*simple walk*) versions. A solution to these problems was found by redesigning the layer structure of Brooks' architecture by moving the balance and compliance layer below walking and other high-level behaviors. According to Porta & Celaya, the resulting organization achieved a clear and natural new subsumption breakdown and solved problems with modularity and incrementality.

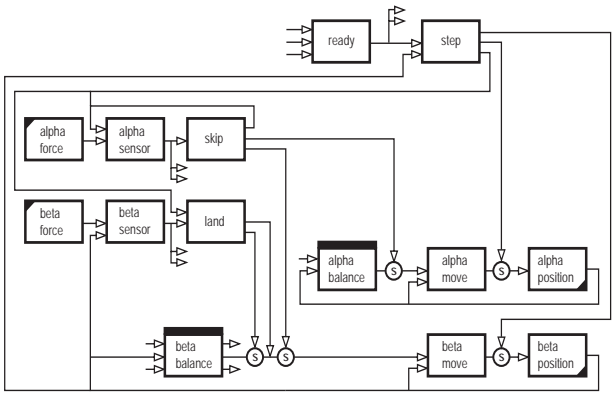

Fig. 5. Subsumption Diagram by Porta & Celaya.

The diagram above represents the subsumption architecture developed by Porta & Celaya. The layout shows that some of the simpler modules, such as *alpha collide* and *up leg trigger* have been encapsulated into the *step* and *skip* modules, and *beta balance* has been moved to a lower level according to the new competence breakdown. The diagram also introduces new modules that are meant to resolve the situation when two or more suppressing signals are connected to the same input. To establish priority levels between these, the motor interface is now split into two behaviors, *alpha move* and *alpha motor*, each with its own supressing signal.

Overall, the new subsumption architecture by Porta & Celaya offers a great improvement over Brooks' in terms of clarity. In particular, the fact that the *beta balance* module is now located at the same low hierarchical level as its counterpart *alpha balance* is a more elegant solution than the high-level placement of this module in Brooks' diagram. As the authors themselves note, "keeping feet and maintaining stability is more fundamental than advancing."

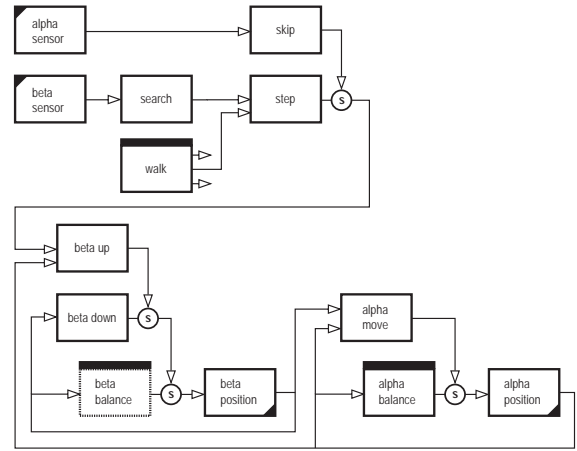

Fig. 6. Our Subsumption Diagram.

## 8. Our Subsumption Architecture

The design approach taken by Porta & Celaya was to start with Brooks' example and to improve upon it by reorganizing the competence layer structure into a more intuitive diagram. However, we feel that Porta & Celaya's architecture could be clarified further. The motor interface modules in particular seem like a peculiar patch that is needed in order to resolve a conflict between higher level machines rather than fulfill a necessary function. This flaw encouraged us to seek a slight modification to the subsumption architecture model for our robot.

The starting point of our subsumption scheme is identical to Brooks' *simple walk*. Next, we add *beta balance* machine at the bottom layer of the diagram. This placement is crucial in re-organization of subsumption modules in that it creates a mirroring of two behavior blocks in the horizontal and vertical direction at the same competence level.

Following this example, our original intention was to add *alpha back* machine above the *alpha forward* module (*alpha advance* in Brooks) as a mirror of the vertical movement machine set. However, there is a difference in which the horizontal and vertical sets operate. The default vertical orientation of the robot is to be standing up, dictating that the *beta down* machines are constantly activated. On the other hand, there is no default position for the horizontal leg movement and thus we see no need for a breakdown of this component in two halves and use a single machine *alpha move* instead.

It is notable that we keep both *beta up* and *beta down* subsumption modules as they provide the basic functionality of moving the robot legs vertically. In contrast, Porta & Celaya retain *beta down* machine (renamed *land*) but eliminate *beta up* as a standalone module, presumably encapsulating its functionality into *skip* and *step* modules. Our architecture contains both of these machines as well, but in comparison to Porta & Celaya's they feature a reduced set of internal states which take care of higher-level functions and communicate with the already existing modules for lower-level tasks.

For example, *skip* contains only the states invoked when a leg encounters an obstacle while moving horizontally. This condition causes *step* to activate *beta up* in order to move the robot leg higher. Accordingly, this module is positioned in the diagram right above and suppresses the output of *step*. Next, the vertical and horizontal force (or touch) sensors are placed on the left side of the diagram. The walking sequence module, which has the internal structure of Brooks' *walk*, completes the layout of our subsumption diagram.

We believe that the resulting layout, though not functionally very different from the model proposed by Porta & Celaya, offers a more clear picture of the system's innerworkings. Perhaps the main advantage is that a slight rearrangement of higher-level modules allows us to avoid the 'ambiguous' condition of two machines (*skip* and *step*) competing to suppress the output of one lower one (*alpha balance*). As a result, the placeholder machines *alpha move* and *beta move* are eliminated.

Our diagram retains the clear visual breakdown of machine blocks into vertical levels of competence, as in Porta & Celaya. In addition, it offers a clear horizontal breakdown of machine blocks into functional categories that could be loosely labeled as horizontal and vertical movement control blocks, higher-level behavior blocks, and sensor input blocks. To sum up, we have attempted to draft a two-dimensional diagram that complements the actual subsumption architecture as a structure that is functional and easy to read.

## 9. Implementation Details

Subsumption architecture is in essence a modular network of finite state machines, each designed to work on a specific behavioral task at hand. IsoPod's programming language IsoMax provides a means to implement this type of architecture using a specially designed syntax. Fig 2. in the earlier section provided an outline of how a finite state machine is constructed in IsoMax. Our subsumption architecture consisted of some 130+ finite state machines programmed in this fashion.

Fig. 7 shows a graph representation of a typical subsumption module in our diagram. The Finite Machine states are represented as circles and transitions between them are indicated by arrows. On the outside of each arrow, a conditional statement is printed which allows the machine to pass from one state to another when fulfilled. On the inside, an action that takes place with a corresponding state change is indicated. The corners of the state machine contain the inputs and outputs of the module that correspond directly to the arrow paths of the subsumption architecture diagram.

The final breakdown of the subsumption architecture module into its representation in programmer's code is shown in Fig 8.

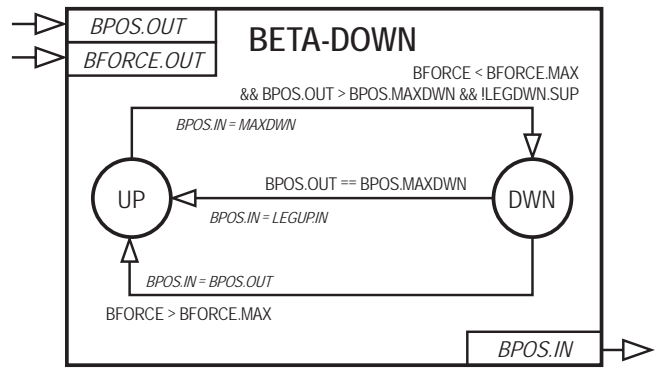

Fig. 7. Graph representation of our *beta down* FSM.

```
MACHINE L1-BETA-DOWN EEWORD
    ON-MACHINE L1-BETA-DOWN
        APPEND-STATE L1-BETA-DOWN-UP EEWORD
        APPEND-STATE L1-BETA-DOWN-DWN EEWORD

    IN-STATE
        L1-BETA-DOWN-UP
    CONDITION
        L1.BPOS.OUT @ BPOS.MAXDWN @ >
        L1.LEGDWN.SUP @ NOT AND
        L1.BFORCE @ < BFORCE.MAX @ AND
    CAUSES
        BPOS.MAXDWN @ L1.BPOS.IN !
    THEN-STATE
        L1-BETA-DOWN-DWN
    TO-HAPPEN IN-EE

    IN-STATE
        L1-BETA-DOWN-DWN
    CONDITION
        L1.BPOS.OUT @ BPOS.MAXDWN @ =
    CAUSES
        L1.BUP.IN @ L1.BPOS.IN !
    THEN-STATE
        L1-BETA-DOWN-UP
    TO-HAPPEN IN-EE
```

Fig. 8. A partial code equivalent of *beta down* FSM.

The code snippet contains the same albeit abbreviated variables, state definitions and conditional statements as the graph diagram above. As it is evident, IsoMax provides a very convenient syntax for defining virtual machines like this, naming the states of the machine and identifying transitions between states. As a result, the relationship between the graph representation and code is very direct and it is literally possible to debug a program by carefully drawing and studying its corresponding diagram.

## 10. Results

IsoPod's Virtually Parallel Machine Architecture appears to provide an excellent platform for implementing a modular control framework for robotics. The goal of our project was to test this hypothesis on a real-world example by programming a subsumption-based behavior architecture.

Our tests were carried out on a pre-fabricated kit robot "Hexapod III", manufactured by Lynxmotion, Inc. in U.S.A. The robot chassis is lasercut from sheets of black lexan and comes with a set of injection molded nuts and bolts. The photo below shows the constructed hexapod, which measures about 10" x 12" x 1.5" and has a ground clearance of about 5 inches. Each of robot's legs has 3 degrees of freedom, and it is powered by 18 Hitec HS-85BB servos. (Lynxmotion, Inc., 2003)

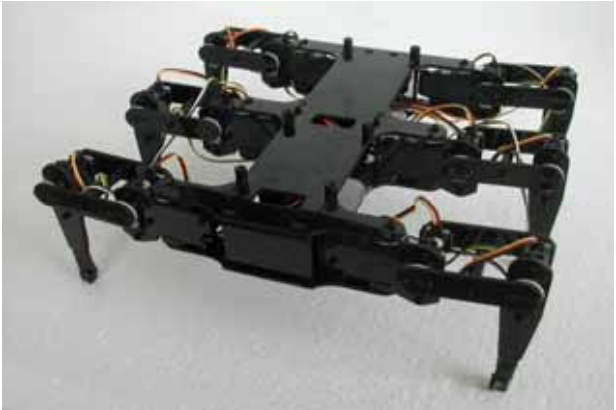

Fig. 9. Hexapod III kit constructed and wired.

The IsoPod board is mounted inside the robot's body and provides direct connections for each servo via a small adapter board. The power supplies for IsoPod and servos are tethered to the robot. A serial connection is used to upload programs to IsoPod hardware using a 115,600kbs connection.

Although our tests of the entire subsumption architecture are still preliminary at this stage, it was sufficient to make the robot walk autonomously using a number of pre-programmed gaits, sense small obstacles and step over them, and to stop and reverse motion at a cliff. Predictably, the main problems we experienced were not with the software architecture but hardware imperfections. For instance, since we used a simple current sensing method of measuring voltage drop across a resistor, contact with ground by force-feedback became difficult to manage (the robot tended to raise itself gradually due to constant *beta down* ouput) and we opted for digital contact sensors on feet instead. Hexapod III also had trouble maintaining its posture occasionally due to the choice of small-size servo motors. Our overall impression, however, was that we were able to design a complex modular control architecture using inexpensive, off-the-shelf parts in a very short time.

## 11. Conclusion

At the time of Brooks' first papers on subsumption, a modular control architecture for a mobile robot could only be implemented with independent processors which sent messages to each other over connecting wires. Each processor in these original designs was itself a finite state machine. For example, the first walking robot built by Brooks had four onboard 8-bit microprocessors linked by 62.5Kbaud token ring, offering a total memory of about 1Kbyte of RAM and 10K of EPROM. (Brooks, R., 1989) The current state of hardware development allows a much greater flexibility in the design and much faster speed of system development. One powerful processor with a large cache of on-board memory and peripherals can replace a whole network of finite state machines and enable an entire subsumption architecture to be constructed in software. The IsoPod/IsoMax development platform is particularly well suited to this task as its architecture is based very directly on the concept of Finite State Machines running in parallel.

Our own preliminary progress presented here demonstrates the feasibility of a quick cycle of development for a complex modular control system of an autonomous robot. Our primary interest has not been to advance the state of technology in robotics, but to suggest that the task of implementing advanced robot control architectures is becoming inexpensive and accessible. We hope that this study will lead to incorporation of such systems into our future projects and encourage more experimentation in this realm by engineers, students and hobbyists.